\pdfoutput=1

\documentclass[11pt]{article}

\usepackage{ACL2023}
\usepackage{graphicx}
\usepackage{subcaption}

\usepackage{times}
\usepackage{latexsym}
\usepackage{tabularx}
\usepackage{amsmath}
\usepackage{amssymb}
\usepackage{algorithm}
\usepackage{algpseudocode}
\usepackage{arydshln}
\usepackage[export]{adjustbox}
\usepackage[T1]{fontenc}

\usepackage[utf8]{inputenc}

\usepackage{microtype}

\usepackage{inconsolata}
\usepackage{xcolor}

\definecolor{myblue}{HTML}{4E84C4}
\definecolor{myred}{HTML}{F56F42}
\definecolor{mygreen}{HTML}{34692E}
\definecolor{myorange}{HTML}{DA7842}

\title{VLEU: a Method for Automatic Evaluation for\\ Generalizability of Text-to-Image Models}

\author{
  Jingtao Cao$^1$, Zheng Zhang$^2$, Hongru Wang$^1$, Kam-Fai Wong$^1$\\
   $^1$The Chinese University of Hong Kong \\
   $^2$Shanghai Jiao Tong University \\
   \texttt{jcao@se.cuhk.edu.hk, kfwang@se.cuhk.edu.hk}
}

\begin{document}
    \maketitle
    \begin{abstract}
    
Progress in Text-to-Image (T2I) models has significantly advanced the generation of images from textual descriptions. Existing metrics, such as CLIP, effectively measure the semantic alignment between single prompts and their corresponding images. However, they fall short in evaluating a model's ability to generalize across a broad spectrum of textual inputs. To address this gap, we propose the VLEU (\underline{\textbf{V}}isual \underline{\textbf{L}}anguage \underline{\textbf{E}}valuation \underline{\textbf{U}}nderstudy) metric. VLEU leverages the power of Large Language Models (LLMs) to sample from the visual text domain, encompassing the entire range of potential inputs for the T2I task, to generate a wide variety of visual text. The images generated by T2I models from these prompts are then assessed for their alignment with the input text using the CLIP model. VLEU quantitatively measures a model's generalizability by computing the Kullback-Leibler (KL) divergence between the visual text marginal distribution and the conditional distribution over the images generated by the model. This provides a comprehensive metric for comparing the overall generalizability of T2I models, beyond single-prompt evaluations, and offers valuable insights during the finetuning process. Our experimental results demonstrate VLEU's effectiveness in evaluating the generalizability of various T2I models, positioning it as an essential metric for future research and development in image synthesis from text prompts. Our code and data will be publicly available at \url{https://github.com/mio7690/VLEU}.
    
    \end{abstract}
    
    \section{Introduction}
    The emergence of latent diffusion models (LDMs) \citep{rombach2022highresolution} marked a significant advancement in generative models, addressing a crucial limitation that was prevalent during the era dominated by Generative Adversarial Networks (GANs) \citep{goodfellow2014generative}. Unlike GANs, which were often constrained by limited expressive and computational capabilities and focused on specific tasks or datasets, LDMs, trained on extensive datasets like LAION-5B \citep{schuhmann2022laion5b}, introduced an enhanced capacity for conditional generation across diverse scenarios. This pivotal development laid the groundwork for major strides in the field of text-to-image (T2I) generation. Notable examples include Stable Diffusion \citep{rombach2022highresolution}, SDXL \citep{podell2023sdxl}, Imagen \citep{saharia2022photorealistic} and DALL-E 3 \citep{openai2023dalle3}, all of which have demonstrated impressive capabilities in generating detailed and contextually relevant images from textual descriptions.

    \begin{figure}[t]
     \includegraphics[width=\columnwidth]{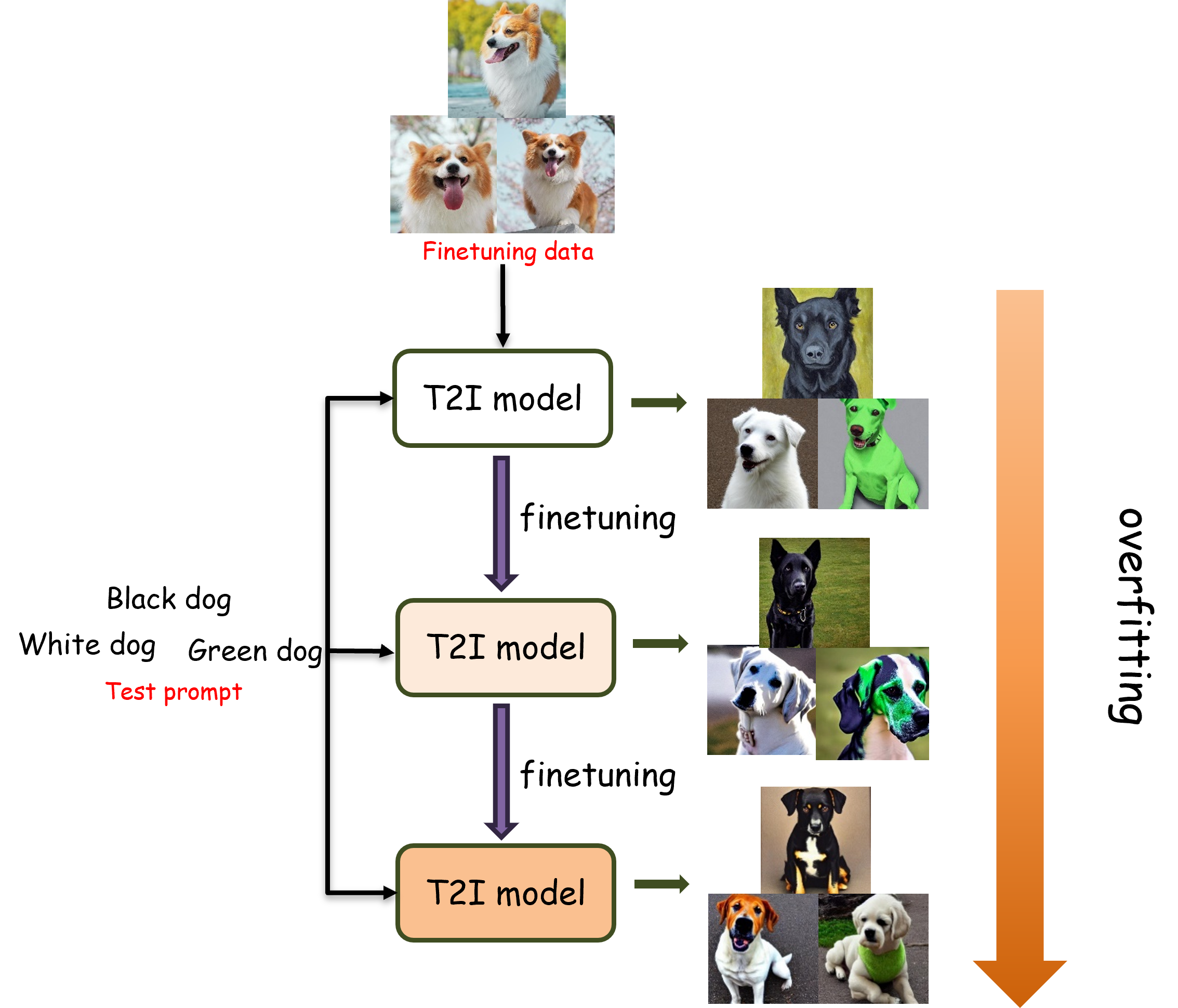}
\caption{\textbf{The loss of generalization of a T2I model.} When fine-tuning a T2I model with images of a brown and white dog, as the fine-tuning process advances, prompts for dogs of various colors start to yield outputs that increasingly reflect the characteristics of the dog present in the training dataset. This results in generated images that deviate from the original textual description, indicating a clear case of \textbf{overfitting} and a \textbf{loss of generalization}.}
     \label{fig:1}
    \end{figure}
    
    When assessing the performance of T2I models, a variety of metrics are employed, including Inception Score (IS) \citep{salimans2016improved}, Fréchet Inception Distance (FID) \citep{heusel2018gans}, which primarily gauge the quality and diversity of the generated images. In contrast, metrics like CLIP score \citep{radford2021learning,hessel2022clipscore}, DINO \citep{caron2021emerging}, and ImageReward \citep{xu2023imagereward} are designed to measure the semantic alignment between the generated images and the input text. Despite the effectiveness of these metrics in their respective domains, they do not fully capture a model's generalizability, which refers to its ability to produce accurate and diverse images across a wide range of textual prompts. This aspect of T2I model performance is often evaluated through subjective human judgment, highlighting the need for a standardized measure of generalizability.
    
    Such issue of evaluating generalizability also extends to the methodologies employed for finetuning T2I models. Techniques such as text inversion \citep{gal2022image}, LoRA \citep{hu2021lora}, DreamBooth \citep{ruiz2023dreambooth}, and HiFi Tuner \citep{wang2023hifi} have been instrumental in adapting pre-trained T2I models to specific subject or style. However, during the evaluation phase of these finetuned models, there is a lack of robust metrics to effectively measure the loss of generalization, which is shown in Figure \ref{fig:1}. Many studies, including DreamBooth among others, have encountered this issue and presented it visually, ultimately relying on subjective human interpretation for assessment.
    
    To bridge this gap, we introduce VLEU (\underline{\textbf{V}}isual \underline{\textbf{L}}anguage \underline{\textbf{E}}valuation \underline{\textbf{U}}nderstudy) metric.\footnote{The name VLEU is inspired by the BLEU metric, which revolutionized the evaluation of machine translation systems \citep{10.3115/1073083.1073135}. The nomenclature draws a parallel insofar as the T2I task can be conceptually likened to translating from text to image.} Firstly we define the input text prompt for the T2I task as visual text, and also the sets of potential inputs for the T2I task as \texttt{visual text domain}. This delineated definition is essential, as not all text is appropriate as input for T2I models. For example, the chat text from the dialogue system, when used as input for T2I models, is nonsensical. VLEU seeks to quantify the generalizability of T2I models by measuring the alignment between the visual text domain and the images generated by the T2I model conditioned on the visual text domain.
    
    VLEU operates by calculating the Kullback-Leibler (KL) divergence \citep{kullback1951information} between the distribution of the visual text domain and the distribution of the images conditionally generated by the model. This divergence serves as a metric for the alignment between the intended text prompts and the generated images. To facilitate this measurement, Large Language Models (LLMs) such as ChatGPT \citep{openai2022chatgpt}, GPT-4 \citep{openai2023gpt4}, and LLaMA \citep{touvron2023llama, touvron2023llama2} are utilized to sample from the visual text domain. These descriptions are then paired with images produced by the T2I model, and the CLIP model \citep{radford2021learning} is used to evaluate the semantic congruence between each text-image pair.
    
    The principal contributions of this work are as follows: 
    \begin{itemize}
        \setlength{\itemsep}{0pt}
        \setlength{\parskip}{0pt}
        \setlength{\parsep}{0pt}
        \item We proposed the VLEU metric, an automatic evaluation designed to assess the generalizability of T2I models.
        \item We detailed the implementation of VLEU, which involves applying LLMs to sample visual text from the visual text domain and utilizing the CLIP model to evaluate the semantic alignment of generated images with the input visual text.
        \item We conducted comprehensive experiments to analyze the effectiveness of VLEU and the impact of different components in the evaluation pipeline, validating its effectiveness in quantifying T2I models' generalizability.
        \item We presented two real-world case studies showcasing the practical utility of VLEU in evaluating T2I models, positioning it as a vital metric for T2I model development.
    \end{itemize}

    \section{Background}
    
    \paragraph{Text-to-Image Generation:} The field of image generation was once dominated by GANs \citep{goodfellow2014generative}, which operate on a framework of competing networks, one generating images and the other evaluating them. However, at this stage, models based on GANs had limited expressive and computational capabilities, which constrained their generalizability across diverse tasks and datasets. For instance, StyleGAN \citep{DBLP:conf/cvpr/KarrasLA19} focused on generating high-quality face images but was not suitable for broader text-to-image generation. GAN research was often focused on specific domains or datasets such as CIFAR-10 \cite{article}, CelebA \citep{liu2015faceattributes} and ImageNet \citep{russakovsky2015imagenet}, rather than being capable of conditional image generation from unconstrained natural language inputs. While there were some promising works exploring conditional generation with GANs model \citep{mirza2014conditional, casanova2021instanceconditioned}, significant challenges remained in bridging the gap between research prototypes and practical applications of text-to-image generation. The pivotal breakthrough came with the introduction of diffusion models, exemplified by DDPM \citep{DBLP:journals/corr/abs-2006-11239} and DDIM \citep{DBLP:journals/corr/abs-2010-02502}. Unlike GANs, diffusion models can synthesize high-fidelity images by gradually denoising random noise through reverse diffusion sampling. This approach provides stronger generative capabilities by leveraging the modeling power of deep neural networks. Furthermore, latent diffusion models like DALL-E \citep{DBLP:journals/corr/abs-2102-12092} and DALL-E 2 \citep{ramesh2022hierarchical} significantly reduced the computational costs of sampling high-resolution images. The release of the LAION-5B dataset \citep{schuhmann2022laion5b}, containing billions of image-text pairs, provided immense amounts of training data to empower text-to-image generation. On top of the latent diffusion framework, the adoption of Transformer architectures\citep{vaswani2017attention} was another vital innovation. Models like Imagen \cite{saharia2022photorealistic} and Stable Diffusion \citep{rombach2022highresolution} incorporated cross-attention layers that align textual prompts with generated image features. This mechanism enabled explicit conditioning on text descriptions and proved crucial for advancing text-to-image capabilities. The breakthrough in scaling up diffusion models to billions of parameters, combined with the effective technique of using Transformer architectures for text conditioning, has led to recent T2I models demonstrating impressive capabilities in synthesizing high-fidelity, controllable images from a wide range of textual descriptions.

    \paragraph{Metrics for T2I Models:} In evaluating T2I models, traditional metrics like Inception Score (IS) \citep{salimans2016improved}, Fréchet Inception Distance (FID) \citep{heusel2018gans} focus on image quality. IS assesses the clarity and diversity of images using a pre-trained Inception network which has fixed classes trained on Imagenet \citep{deng2009imagenet}, while FID measures the distance between feature vectors of real and generated images to gauge realism. Meanwhile, newer metrics such as DINO \citep{caron2021emerging}, CLIP similarity \citep{radford2021learning,hessel2022clipscore}, and ImageReward \citep{xu2023imagereward} offer a more nuanced assessment. DINO focuses on discerning the similarity between generated and actual images by emphasizing distinctive features. CLIP similarity metric examines the congruence between images and their textual descriptions. ImageReward gauges the aesthetic and creative attributes of generated images, aligning them with human aesthetic preferences. However, all these metrics primarily concentrate on either image quality or semantic alignment, revealing a gap in evaluating a model's generalizability across varied textual prompts.
    
    \paragraph{Large Language Models:} The landscape of natural language processing has been revolutionized by the advent of LLMs like ChatGPT \citep{openai2022chatgpt}, GPT-4 \citep{openai2023gpt4}, and LLaMA \citep{touvron2023llama, touvron2023llama2}. These models exhibit remarkable abilities in understanding context, generating coherent and contextually relevant text, and mimicking human conversational styles. These LLMs are not only adept in producing context-aware and coherent text but are also effectively utilized to generate prompts that guide text-to-image models. For example, recent work Visual ChatGPT \citep{wu2023visual} demonstrates using ChatGPT to automatically generate prompts for text-to-image models. The conversational nature of ChatGPT allows generating prompts with greater contextual awareness and abstraction compared to human-written prompts.

    \section{VLEU}

    \begin{figure}
     \centering
     \includegraphics[width=\columnwidth]{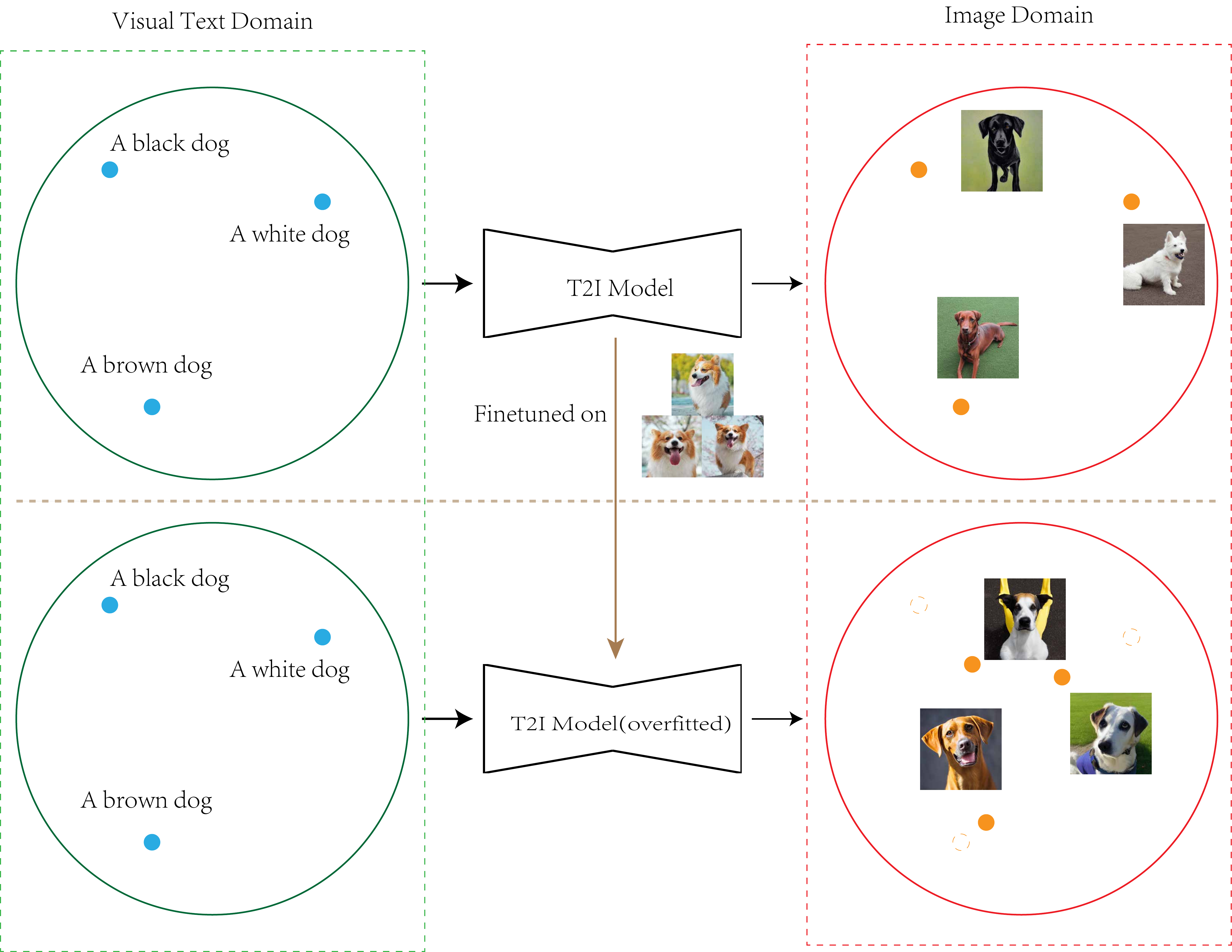}
    \caption{\textbf{Generalizability in T2I Models: A Comparative Visualization.} The first model demonstrates strong generalization by successfully producing images that align with various input prompts. In contrast, the second model shows overfitting to the prompt "A white and yellow dog," leading to a failure to generalize to other inputs. As a result, its generated images are generally misaligned with the given prompts. Our proposed VLEU metric aims to quantify this observation.}
     \label{fig:vtd}
    \end{figure}

    \begin{table*}
    \centering
    \begin{tabularx}{\textwidth}{>{\hsize=.5\hsize}X>{\hsize=1.5\hsize}X}
    \hline
    \textbf{Prompt Type} & \textbf{Example Template} \\
    \hline
    \textcolor{mygreen}{Unconstrained} \textcolor{myorange}{Subject} & Please imagine a random picture and describe it in one sentence. \\
    \textcolor{mygreen}{Constrained} \textcolor{myorange}{Subject} & Please imagine a picture of \{\textcolor{myred}{class\_word}\} and describe it in one sentence, making sure to include the word "\{\textcolor{myred}{class\_word}\}". \\
    \textcolor{mygreen}{Constrained} \textcolor{myorange}{Subject} \newline with \textcolor{myorange}{Properties} & Please imagine a picture of \{\textcolor{myred}{class\_word}\} and describe it in one sentence, making sure to include the word "\{\textcolor{myred}{class\_word}\}" and words about \newline \{\textcolor{myred}{property}\}. \\
    \hline
    \end{tabularx}
    \caption{\textbf{Templates used for sampling T2I prompts.} These templates serve as inputs for chat-based LLMs, such as ChatGPT, to generate T2I prompts. LLMs leverage their understanding of natural language to produce diverse and contextually relevant prompts within the visual text domain.}
    \label{table:prompts}
    \end{table*}

    How can we quantify the generalizability of a T2I model? This seems like an intractable problem at first glance. Our intuition stems from observing the phenomenon of T2I models losing generalizability during finetuning.
    
    As shown in Figure \ref{fig:vtd}, through comparison, we can find that if the T2I model can generate good images for all prompts, then the generated images should be aligned with the given prompts overall. If the model loses the ability to generate some prompts, it will cause an overall misalignment. We quantify this alignment as the KL divergence between the visual text marginal distribution and the conditional distribution over the images generated by the model, which VLEU aims to measure.
    
    VLEU employs a three-step automatic process to measure the above KL divergence. First, text prompts are sampled from the visual text domain and used to generate corresponding images (detailed in Section \ref{visual_text_sampling}). Second, the CLIP model evaluates the semantic alignment between each generated image and its original textual prompt (explained in Section \ref{text_image_scoring}). Finally, these text-image alignments are utilized for probability modeling to obtain the visual text marginal distribution and the conditional distribution over generated images. These distributions are then used to compute the final VLEU score (formulation provided in Section \ref{subsec:vleu-calculation}).

    \subsection{Visual Text Sampling}
    \label{visual_text_sampling}
    To initiate the evaluation process, we sample from the visual text domain - the space of potential textual inputs for the T2I task. To effectively sample from the expansive visual text domain without requiring extensive manual effort, we leverage the generalizability of LLMs to automatically generate diverse T2I prompts that closely approximate the broad sampling from the visual text domain. In this approach, we utilize LLMs to sample two types of prompts: unconstrained subject and constrained subject.
    
    Table \ref{table:prompts} demonstrates the templates used for sampling T2I prompts. The prompt template for constrained subjects is tailored to ensure that the generated prompts contain the same class word. This consistency is vital when evaluating the loss of generalization in relation to a specific word. For instance, if “dog” is the class word, but the LLM replaces it with synonyms like “pooch”, “hound”, or “pup” in the T2I prompts, it could obscure the true extent to which the model's generalizability to the word “dog” has been affected.
    \begin{figure*}
        \centering
        \begin{subfigure}{0.25\textwidth}
            \includegraphics[width=\linewidth]{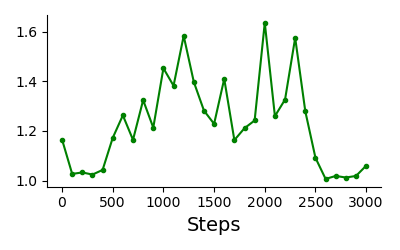}
            \subcaption{IS}
        \end{subfigure}%
        \begin{subfigure}{0.25\textwidth}
            \includegraphics[width=\linewidth]{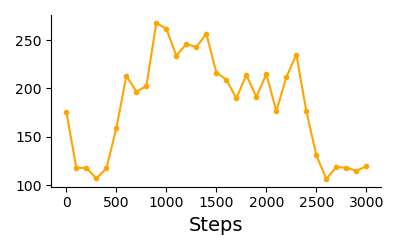}
            \subcaption{FID}
        \end{subfigure}%
        \begin{subfigure}{0.25\textwidth}
            \includegraphics[width=\linewidth]{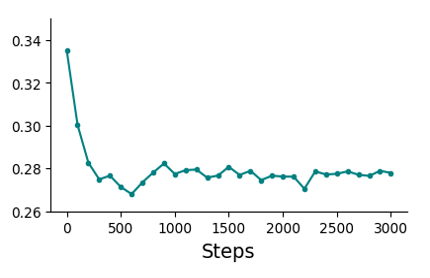}
            \subcaption{CLIP (avg.)}
        \end{subfigure}%
        \begin{subfigure}{0.25\textwidth}
            \includegraphics[width=\linewidth]{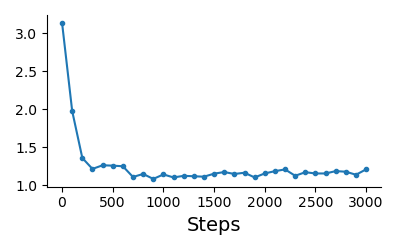}
            \subcaption{VLEU}
        \end{subfigure}
        \caption{\textbf{Variation of different metrics during finetuning.} In this example, we finetuned SD 1.5 on 5 specific teddy bear images and sampled 25 prompts from the visual text domain covered by various teddy bears using ChatGPT 3.5. For FID, we treated the images in the dataset as the real image distribution.}
        \label{fig:ifv}
    \end{figure*}

    Additionally, despite increasing the diversity of output by adjusting parameters like the temperature in the ChatGPT API, the responses can still exhibit a degree of convergence. To counteract this and further diversify the T2I prompts, a multi-turn dialogue approach is adopted. After the initial use of the above-mentioned templates, subsequent interactions simply use the prompt “Again” to stimulate the generation of new T2I prompts. This dialogue can span up to 50 rounds, excluding the more convergent prompts typically produced in the first round.
    
    \subsection{Text-Image Scoring}
    \label{text_image_scoring}
    Using the T2I prompts sampled from the visual text domain, corresponding images are generated by the test model under evaluation. To assess the semantic alignment between each input prompt and output image, we leverage CLIP (Contrastive Language-Image Pre-training)\citep{radford2021learning}, which has demonstrated strong capabilities in matching textual descriptions to images.
    
    Specifically, we obtain embedded representations of the text prompts and generated images from CLIP. The similarity between the text and image embeddings for each pair is then quantified using cosine similarity. Cosine similarity measures the angle between two vectors in high-dimensional space, providing a bounded similarity score that is robust to distortions from large vector magnitudes. This enables an effective assessment of how well the generated image aligns with the semantic concepts expressed in the original textual prompt.

    \subsection{VLEU Calculation}
    \label{subsec:vleu-calculation}
    
    The Visual Text Sampling and Text-Image Scoring modules together provide the foundation for the VLEU metric. The VLEU metric quantitatively measures the generalizability of a T2I model by computing the divergence between two probability distributions - the marginal visual text distribution $P(x)$, and the conditional distribution $P(x|y)$ over text prompts given a generated image.
    
    Let $\mathbf{G}$ denote the T2I model being evaluated. Given a corpus $X = {x_1, x_2, ..., x_N}$ of $N$ textual prompts sampled from the visual text domain, the model generates corresponding image $y_i = \mathbf{G}(x_i)$. The similarity between each text-image pair $(x_i, y_j)$ is scored using CLIP as $S_{ij}$.

    These similarity scores are transformed into a conditional distribution over text prompts associated with each generated image via a softmax function:
    \begin{equation}
    P(x|y_i) = \textrm{softmax}(S_{:,i}/t) \label{eq:softmax}
    \end{equation}
    where $t$ is a temperature parameter. The marginal distribution is obtained by averaging over the conditionals:
    \begin{equation}
    P(x_i) = \frac{1}{N}\sum_{y} P(x_i|y)
    \end{equation}
    
    Finally, the VLEU score is computed as the exponentiated expected KL divergence between the conditional and marginal distributions:
    \begin{equation}
    \textrm{VLEU} = \exp\left(\mathbb{E}_{x}\left[\textrm{KL}(P(x|y) | P(x))\right]\right)
    \end{equation}
    Taking the exponentiation is to scale the scores into a more convenient range for comparison. This provides an interpretable measure of the model's ability to generate diverse images aligned with the visual text domain.

    \begin{figure*}
     \centering
     \includegraphics[width=\textwidth]{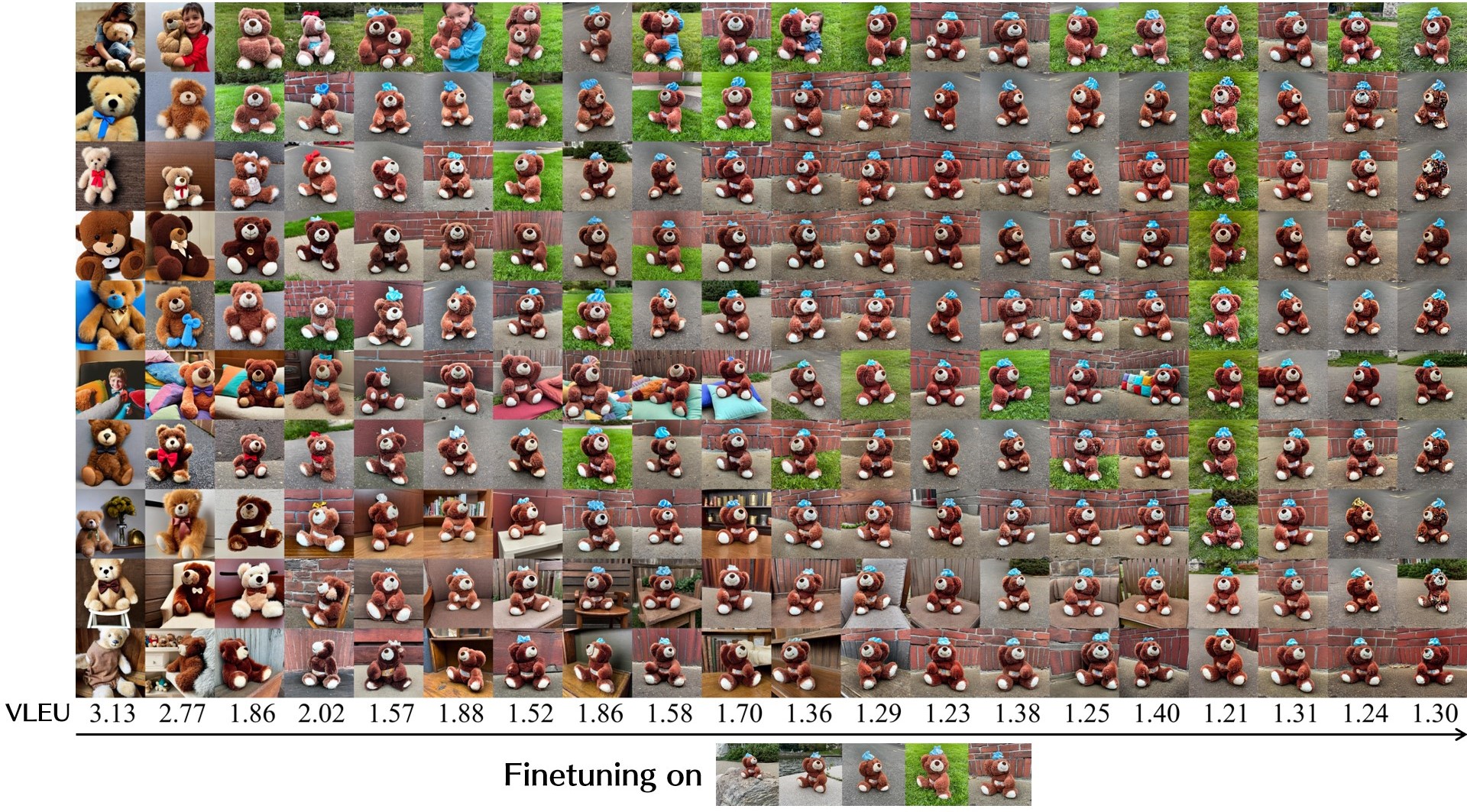}
     \caption{\textbf{Changes in VLEU of a T2I model during finetuning.} Throughout the finetuning of SD 1.5 on five particular teddy bear images, images are generated at every 20 steps using the same prompts. Meanwhile, we calculate the VLEU score at the same step. The figure indicates that the VLEU score gradually decreases as the model begins to overfit, resulting in a loss of generalization and the values align well with this trend.}
     \label{fig:grad}
    \end{figure*}

    \section{Experiments And Analysis}
    
    This section investigates the effectiveness of VLEU in evaluating T2I models and analyzes the impact of different components within the evaluation pipeline. Specifically, we conducted two experiments:
    (1) Analyze VLEU's effectiveness in capturing model generalizability changes during finetuning and across different T2I models, which is presented in section \ref{exp1}.
    (2) Assess the impact of key components like Visual Text Sampler and Text-Image Scorer on VLEU scores, which is detailed in section \ref{exp2}.

    In line with our objectives, we primarily evaluated four open-source T2I models (SD 1.5, SD 2.0, SD 2.1, SDXL) under various conditions. For finetuning, we utilized the dataset provided by DreamBooth \citep{ruiz2023dreambooth}, which comprises several subsets, each containing a series of images related to a specific subject.\footnote{The dataset can be found at \texttt{\href{https://github.com/google/dreambooth}{google/dreambooth}}.} Besides, within all VLEU calculations, the value of temperature in Equation \eqref{eq:softmax} was set to 0.01 to scale the computed results into a visually convenient range for analysis purposes.

    \subsection{Effectiveness Analysis}
    \label{exp1}

    \paragraph{Analysis on Finetuning.} We selected the finetuning process of T2I models on specific datasets, which is often considered detrimental to the models' generalizability, and tested the changes of IS \citep{salimans2016improved}, FID \citep{heusel2018gans}, CLIP score  and VLEU during finetuning.\footnote{For the computation of IS and FID, we utilized \texttt{\href{https://github.com/toshas/torch-fidelity}{toshas/torch-fidelity}}.} Specifically, we finetuned SD 1.5 on the DreamBooth dataset and sampled 25 prompts related to each subset using GPT 3.5. For FID calculation, we treated the images in the dataset as the real image distribution. Note that the FID results may be affected by the small size of the image set. For CLIP score, we simply calculate the average clip similarity of all prompts and corresponding image.

    Taking the finetuning on teddy bear images as an example, as shown in Figure \ref{fig:grad}, the generated images during the finetuning process tend to increasingly resemble images from the training set as the number of finetuning steps increases, indicating a decrease in model generalizability. Figure \ref{fig:ifv} illustrates the trends of IS, FID, CLIP score and VLEU throughout the finetuning process. Among these, VLEU consistently decreases, aligning more closely with the diminishing generalizability during finetuning compared to the CLIP score. However, IS and FID, which are commonly used metrics reflecting image clarity and diversity, exhibited significant fluctuations. This discrepancy can be attributed to the fact that while diversity consistently decreases during finetuning, changes in image clarity may not follow a consistent pattern.

    \begin{table*}
    \centering
    \makebox[\textwidth][c]{
    \small
    \begin{tabular}{l:ccc:ccc:ccc:ccc}
    \hline
    \textbf{Models} & \multicolumn{3}{c:}{\textbf{Unconstrained}} & \multicolumn{3}{c:}{\textbf{Scene-Focused}} & \multicolumn{3}{c:}{\textbf{Person-Focused}} & \multicolumn{3}{c}{\textbf{Style-Focused}} \\
    & \textbf{VLEU} & \textbf{Elo} & \textbf{CLIP} & \textbf{VLEU} & \textbf{Elo} & \textbf{CLIP} & \textbf{VLEU} & \textbf{Elo} & \textbf{CLIP} & \textbf{VLEU} & \textbf{Elo} & \textbf{CLIP} \\
    \hline
    SD 1.5 & 37.18 & 934 & 0.3254 & 43.96 & 921  & 0.3329 & 48.79 & 894 & 0.3029 & 58.78 & 875 & 0.3355 \\
    SD 2.0 & 41.78 & 1048 & 0.3251 & 49.46 & 1011 & 0.3320 & \textbf{50.55} & \textbf{1263} & 0.3044 & \textbf{59.65} & 1087 & 0.3360 \\
    SD 2.1 & \textbf{42.05} & \textbf{1101} & 0.3264 & \textbf{50.63} & \textbf{1086} & 0.3330 & 50.04 & 980 & 0.3051 & 59.14 & \textbf{1111} & 0.3367 \\
    SDXL & 39.15 & 917 & \textbf{0.3297} & 48.60 & 983 & \textbf{0.3352} & 45.68 & 863  & \textbf{0.3055} & 55.97 & 927 & \textbf{0.3381} \\
    \hline
    \end{tabular}
    }
    \caption{\textbf{The comparison between VLEU, CLIP, and Elo scores for different models.} We tested several common T2I models on VLEU scores across the entire visual text domain, as well as within the subdomains of scene, person, and style. We sampled 1000 prompts in each domain using ChatGPT 3.5. Additionally, we conducted a human evaluation study to compute Elo ratings for each model, based on pairwise comparisons of images generated from human-created prompts. \textbf{CLIP} refers to the average of CLIP scores obtained from text-image scoring, not using our VLEU calculation.}
    \label{table:t2im}
    \end{table*}

    \paragraph{Analysis across T2I models.} We applied our VLEU metrics to comprehensively evaluate four open-source T2I models across four major visual text domains, including unconstrained, scene-focused, person-focused, and style-focused visual text. We sampled 1000 prompts in each domain using GPT 3.5. Additionally, we calculated CLIP scores, which are obtained by simply averaging CLIP similarity scores without using our VLEU calculation.
    
    As a comparison, we also conducted a human evaluation study, where human evaluators were asked to create a variety of T2I prompts, either freely or focused on a specific subject, similar to how LLMs generate prompts. These prompts were then used to generate images with the four T2I models. The evaluators compared pairs of images generated by randomly selected pairs of models, without knowing which model produced which image, and determine which image better adhered to the prompts. We used these pairwise comparisons to compute an Elo rating \citep{elo1967uscf} for each model, which is a method commonly used to calculate the relative levels of players in win-loss games. 
    
    As shown in Table \ref{table:t2im}, the Elo ratings derived from human evaluations were consistent with the VLEU scores, with SD 2.0 and SD 2.1 outperforming SD 1.5 and SDXL across most domains. This alignment between human evaluation and VLEU scores
    
    demonstrates the effectiveness of our VLEU metric in assessing the generalizability of T2I models, and proves that our VLEU calculation reflects the generalizability better than simply averaging CLIP similarity scores.
    
    For a detailed explanation of the Elo rating calculation and the human evaluation process, please refer to the Appendix \ref{sec:human}.

    \subsection{Component Impact Analysis}
    \label{exp2}
    In the VLEU pipeline, we focused on two key components: the Visual Text Sampler and the Text-Image Scorer. These components play pivotal roles in shaping the effectiveness and outcomes of the VLEU system. In this analysis, we investigated the impact of utilizing different models for these components on the efficacy of the VLEU pipeline.
    
    \paragraph{Visual Text Sampler.} We experimented with four widely used LLMs (GPT-3.5-turbo, GPT-4, LLaMA-2-7B-Chat, LLaMA-2-13B-Chat \citep{touvron2023llama2}) as the Visual Text Sampler within the VLEU pipeline and assessed their effectiveness. For each sampler, We sampled 1000 subject-unconstrained prompts and evaluated them on different T2I models. It's worth noting that because the two LLaMA models have a weaker ability to follow instructions, they couldn't directly output T2I prompts in the expected format under zero-shot conditions. Hence, we manually crafted initial dialogues for the first two rounds in a few-shot manner to guide the model in generating T2I prompts in the desired format. As depicted in Table \ref{table:llm}, while different samplers exert varying influences on VLEU scores, the overall ranking of several T2I models remains largely consistent. Additionally, we observed that utilizing LLMs deemed to have better generalizability, such as GPT-4, as the Visual Text Sampler resulted in higher VLEU scores compared to those obtained with LLMs with poorer generalizability, such as LLaMA-2-7B. We hypothesize that this phenomenon stems from the weaker descriptive capability of prompts generated by LLMs with poor generalizability, leading to substantially low similarity between the generated images and these prompts. Consequently, this increases the KL divergence of each image relative to the marginal distribution for text embeddings, thereby yielding higher VLEU scores.

    \begin{table*}
    \centering
    \begin{tabular}{lcccc}
    \hline
    \textbf{T2I Model} & \textbf{ChatGPT 3.5} & \textbf{GPT-4} & \textbf{LLaMA-2-7B} & \textbf{LLaMA-2-13B} \\
    \hline
    SD 1.5 & 37.18 & 36.85 & 58.89 & 47.79 \\
    SD 2.0 & 41.78 & 40.72 & \textbf{60.15} & 49.27 \\
    SD 2.1 & \textbf{42.05} & \textbf{41.48} & 59.12 & \textbf{49.41} \\
    SDXL & 39.15 & 39.33 & 58.30 & 48.27 \\
    \hline
    \end{tabular}
    \caption{\textbf{VLEU scores under different Visual Text Samplers.} For each sampler, We sampled 1000 subject-unconstrained prompts and evaluated them on different T2I models.}
    \label{table:llm}
    \end{table*}
    
    \begin{table*}
    \centering
    \begin{tabular}{lcccc}
    \hline
    \textbf{T2I Model} & \textbf{CLIP-ViT-B-16} & \textbf{CLIP-ViT-L-14} & \textbf{OpenCLIP-ViT-L-14} & \textbf{OpenCLIP-ViT-H-14} \\
    \hline
    SD 1.5 & 37.18 & 45.86 & 88.82 & 98.83 \\
    SD 2.0 & 41.78 & 51.01 & 102.45 & \textbf{133.76} \\
    SD 2.1 & \textbf{42.05} & \textbf{52.64} & \textbf{104.22} & 132.72 \\
    SDXL & 39.15 & 50.87 & 96.04 & 111.56 \\
    \hline
    \end{tabular}
    \caption{\textbf{VLEU scores using different Text-Image Scorer.} We retained 1000 subject-unconstrained prompts from ChatGPT 3.5 along with corresponding images generated by each T2I model, only changing the Text-Image Scorer in the pipeline to compute the final VLEU scores.}
    \label{table:clip}
    \end{table*}

    \paragraph{Text-Image Scorer.} We explored four different Text-Image Scorers (CLIP-ViT-B-16, CLIP-ViT-L-14 \citep{radford2021learning}, OpenCLIP-ViT-L-14, OpenCLIP-ViT-H-14 \citep{cherti2023reproducible}) for computing VLEU scores. We retained 1000 subject-unconstrained prompts from GPT 3.5 along with corresponding images generated by each T2I model, only changing the Text-Image Scorer in the pipeline to compute the final VLEU scores. As shown in Table \ref{table:clip}, it can be observed that the higher-performing scorers yield higher VLEU scores. We attribute this to the superior matching capability of the higher-performing scorers in aligning images with textual prompts. Consequently, each image exhibits greater discrepancies in scores between its own prompt and other prompts, resulting in larger KL divergences computed and thus higher final scores.
    
    \begin{table}
    \centering
    \begin{tabular}{lcccc}
    \hline
    \textbf{T2I Model} & \textbf{African} & \textbf{Asian} & \textbf{Caucasian} \\
    \hline
    SD 1.5 & 34.45 & 25.89 & \underline{89.96} \\
    SD 2.0 & 36.59 & \textbf{29.50} & \underline{92.99} \\
    SD 2.1 & \textbf{38.03} & 28.13 & \underline{\textbf{95.26}} \\
    SDXL & 32.88 & 22.04 & \underline{92.93} \\
    \hline
    \end{tabular}
    \caption{\textbf{VLEU scores across different ethnicities.} We sampled 1000 prompts from the visual text domain of each ethnicity using ChatGPT 3.5 and computed VLEU for several common T2I models on these prompts. Bold highlights the best score among all models, and underline underscores the best score across three races.}
    \label{table:race}
    \end{table}
    
    \section{Case Studies Using VLEU}
    
    In this section, we present two case studies to illustrate the practical application of VLEU.
    
    \paragraph{Racial Bias in T2I Models.} We tested four T2I models to evaluate their VLEU scores across African, Asian, and Caucasian people. For each ethnicity, we sampled 1000 prompts using GPT 3.5\footnote{The prompts are provided in the supplementary materials}. The results, displayed in Table \ref{table:race}, indicate that all tested models achieved higher scores on Caucasians compared to Africans and Asians. This suggests that these models exhibit higher generalizability performance on Caucasians, due to significant disparities in the representation of different racial groups within the training data.
    
\begin{figure}
     \centering
     \includegraphics[width=\columnwidth]{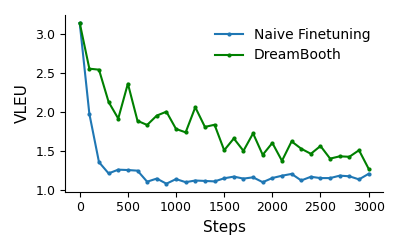}
     \caption{\textbf{VLEU of different finetuning methods.} We finetuned SD 1.5 on 5 specific teddy bears. For DreamBooth, we used 25 teddy bear images generated by the initial model as class images. During the evaluation of VLEU, we used 25 prompts about teddy bears sampled by ChatGPT 3.5.}
     \label{fig:db}
    \end{figure}

    \paragraph{Finetuning Methods Comparison.} We compared the performance of two finetuning methods, naive finetuning and Dreambooth, on the SD 1.5 model using the DreamBooth dataset. Specifically, we select a subset of 5 teddy bear images as an example. We sampled 25 prompts about teddy bears and calculated the VLEU scores during finetuning. As expected and shown in Figure \ref{fig:db}, Dreambooth showed a slower decline in VLEU compared to naive finetuning, aligning with its goal of preserving model generalizability during specialized training. The discernible gap in VLEU curves validates its sensitivity in capturing different rates of generalization loss. This demonstrates VLEU's efficacy in evaluating finetuning methods' ability to balance specificity and generalizability, a valuable asset for model development.
    
    \section{Conclusion}
    
    We introduced VLEU, a novel automatic metric to evaluate T2I models' generalizability. VLEU quantifies alignment between sampled visual text prompts and generated images using LLMs and CLIP. Through experiments and case studies, we demonstrated VLEU's efficacy in capturing declining generalizability during finetuning, discerning differences across models, and comparing finetuning techniques. VLEU provides an automated, standardized metric accounting for a broad space of textual prompts. Our results validate VLEU as an effective metric for quantifying T2I models' generalizability.
    
    \section{Limitations}

    The efficacy of VLEU is constrained by the expressive capabilities of LLMs as the visual text sampler. Even state-of-the-art models struggle to achieve full coverage of the expansive visual text domain when sampling prompts, limiting prompt diversity for evaluation. As language models advance further in natural language understanding and generation, the accuracy and robustness of the VLEU metric will also improve.
    
    Determining sufficient prompt quantities for robust evaluation also presents a challenge. Our experiments indicate this depends on the application scenario. For tracking declining generalizability during finetuning, relatively small samples (around 25 prompts) suffice. However, comparing generalizability across models necessitates larger samples (e.g. 1000 prompts) to ensure evaluation rigor. Further research could systematically investigate optimal prompt quantities for varying contexts to enhance VLEU stability.
    
    While limitations exist, VLEU remains a promising metric providing a standardized framework for evaluating and improving T2I models' generalizability. Future work can explore sampling strategies and evaluation configurations to enhance VLEU robustness and utility.

\section{Ethical Considerations}

Our research adheres to stringent ethical standards. We utilized publicly available datasets that have been ethically vetted to avoid offensive or biased content. Participants in our human evaluation were fairly compensated, ensuring ethical treatment. Consequently, our study presents no ethical concerns, as the data is ethically sourced, the analysis unbiased, and all procedures comply with established ethical guidelines.

\section{Acknowledgement}
This research work is partially supported by ITF (No. PRP/054/21FX), CUHK direct grant (No. 4055209), and CUHK Knowledge Transfer Project Fund (No. KPF23GWP20).

\bibliography{anthology,custom}
\bibliographystyle{acl_natbib}

\clearpage
\appendix

\begin{table*}
\centering
\begin{subtable}[t]{\columnwidth}
\centering
\begin{tabular}{lcccc}
\hline
\textbf{N prompts} & \textbf{SD 1.5} & \textbf{SD 2.0} & \textbf{SD 2.1} & \textbf{SDXL} \\
\hline
 100 &       21.54 &       22.23 &       22.43 &       \textbf{23.33} \\
 100 &       13.89 &       13.88 &       \textbf{15.08} &       14.63 \\
 200 &       21.98 &       22.26 &       23.46 &       \textbf{23.54} \\
 200 &       26.71 &       28.58 &       29.39 &       \textbf{30.87} \\
 300 &       25.61 &       26.60 &       28.25 &       \textbf{29.13} \\
 300 &       34.72 &       36.46 &       \textbf{38.65} &       38.29 \\
 400 &       30.71 &       32.08 &       33.49 &       \textbf{34.55} \\
 400 &       35.71 &       36.44 &       \textbf{38.29} &       37.78 \\
 500 &       35.16 &       36.72 &       38.62 &       \textbf{39.37} \\
 500 &       31.76 &       33.73 &       34.81 &       \textbf{35.41} \\
1000 &       37.18 &       39.15 &       41.78 &       \textbf{42.05} \\
1000 &       29.17 &       30.12 &       32.82 &       \textbf{32.87} \\
\hline
\end{tabular}
\caption{Unconstrained subject}
\end{subtable}
\quad
\begin{subtable}[t]{\columnwidth}
\centering
\begin{tabular}{lcccc}
\hline
\textbf{N prompts} & \textbf{SD 1.5} & \textbf{SD 2.0} & \textbf{SD 2.1} & \textbf{SDXL} \\
\hline
 100 &       20.33 &       \textbf{20.41} &       20.33 &       18.97 \\
 100 &       21.66 &       \textbf{23.51} &       19.71 &       19.15 \\
 200 &       32.18 &       \textbf{33.37} &       30.79 &       28.71 \\
 200 &       33.76 &       33.20 &       \textbf{34.31} &       30.76 \\
 300 &       38.96 &       39.33 &       \textbf{39.40} &       34.14 \\
 300 &       36.05 &       34.46 &       \textbf{36.59} &       32.87 \\
 400 &       43.67 &       \textbf{43.97} &       43.40 &       38.85 \\
 400 &       37.13 &       36.38 &       \textbf{37.50} &       34.27 \\
 500 &       46.19 &       \textbf{46.82} &       46.03 &       41.91 \\
 500 &       38.04 &       38.61 &       \textbf{39.11} &       36.32 \\
1000 &       48.79 &       \textbf{50.55} &       50.04 &       45.68 \\
1000 &       28.12 &       29.31 &       \textbf{29.92} &       27.47 \\
\hline
\end{tabular}
\caption{Person-focused}
\end{subtable}
\caption{\textbf{VLEU scores for different sample sizes.} Each row represents an independent sampling. The two rows with the same sample size represent two distinct sampling instances under identical conditions to investigate whether the relative order of VLEU calculation results remains stable for the same sample size. We primarily sampled two types of prompts for experimentation: unconstrained subject and person-focused prompts.}
\label{table:num}
\end{table*}

\section{Implementation Details}

In this section, we provide a detailed description of the implementation process for our proposed VLEU metric. The implementation consists of two main components: sampling text prompts and calculating the VLEU score. We provide the pseudocode for each component to facilitate understanding and reproducibility.

\subsection{Sampling Text Prompts}

The first step in our process is to sample text prompts from the visual text domain. We use LLMs to generate these prompts. The prompts can either be random or contain a specific keyword. The pseudocode described in Algorithm \ref{alg:stp} outlines the process of generating text prompts.

\begin{algorithm*}
\caption{Sampling Text Prompts}
\label{alg:stp}
\begin{algorithmic}[1]
\Require number of prompts $num$, keyword $key\_word$ (optional), include keyword $include\_key\_word$ (optional), step size $step$
\Ensure A list of text prompts
\State Initialize LLM
\State Initialize an empty list $prompts$

\For{$i \gets 0$ to $num$ by $step$}
    \If{$key\_word$ is not None}
        \If{$include\_key\_word$}
            \State $system\_input \gets$ \texttt{`Please imagine a picture of \{\$key\_word\} and describe it in one sentence, making sure to include the word "\{\$key\_word\}."'}
        \Else
            \State $system\_input \gets$ \texttt{`Please imagine a picture of random \{\$key\_word\} and describe it in one sentence.'}
        \EndIf
    \Else
        \State $system\_input \gets$ \texttt{`Please imagine a random picture and describe it in one sentence.'}
    \EndIf
    \State $human\_input \gets$ [\texttt{SystemMessage(content=system\_input)}]
    \State $ai\_output \gets$ \texttt{llm(human\_input)}
    \State $n \gets 0$
    \State $limit \gets \min(step, num - i)$
    \While{$n < limit$}
        \State Append \texttt{AIMessage(content=ai\_output.content)} to $human\_input$
        \State Append \texttt{HumanMessage(content='Again')} to $human\_input$
        \State $ai\_output \gets$ \texttt{llm(human\_input)}
        \While{$key\_word$ is not None and $include\_key\_word$ and $key\_word$ not in $ai\_output.content$}
            \State $ai\_output \gets$ \texttt{llm(human\_input)}
        \EndWhile
        \State Append $ai\_output.content$ to $prompts$
        \State Print $ai\_output.content$
        \State $n \gets n + 1$
    \EndWhile
\EndFor
\State Return $prompts$
\end{algorithmic}
\end{algorithm*}

\subsection{Calculating the VLEU Score}

The second step involves calculating the VLEU score using the CLIP model to evaluate the semantic alignment between generated images and their corresponding text prompts. The pseudocode described in Algorithm \ref{alg:cvs} outlines the process of calculating the VLEU score.

\begin{algorithm*}
\caption{Calculating the VLEU Score}
\label{alg:cvs}
\begin{algorithmic}[1]
\Require T2I model $t2i\_model$, CLIP model $model$, T2I prompts $prompts$, number of prompts $num$, temperature $T$
\Ensure VLEU score for $t2i\_model$
\State Load CLIP model and processor from $model$
\State Load prompts from $prompts\_path$
\For{$m \in model\_types$}
    \State Initialize $text\_embs \gets []$, $img\_embs \gets []$
    \For{$p \in prompts$}
        \State Tokenize $p$, move to device
        \State Get and normalize text features from CLIP
        \State Append to $text\_embs$
    \EndFor
    \For{$p \in prompts$}
        \State Generate image $img$ from prompt $p$ using $t2i\_model$
        \State Move $img$ to device, preprocess if necessary
        \State Get and normalize image features from CLIP
        \State Append to $img\_embs$
    \EndFor
    \State Initialize $prob\_matrix \gets []$
    \For{$img\_emb \in img\_embs$}
        \State Initialize $cosine\_sim \gets []$
        \For{$text\_emb \in text\_embs$}
            \State Calculate cosine similarity between $img\_emb$ and $text\_emb$
            \State Append to $cosine\_sim$
        \EndFor
        \State Calculate probability distribution using softmax with $T$
        \State Append to $prob\_matrix$
    \EndFor
    \State Stack $prob\_matrix$ into tensor
    \State Calculate marginal distribution for text embeddings
    \State Initialize $image\_kl\_divergences \gets []$
    \For{$prob \in prob\_matrix$}
        \State Calculate KL divergence for current image
        \State Append to $image\_kl\_divergences$
    \EndFor
    \State Calculate VLEU score as $\exp(\text{avg KL divergence})$
\EndFor
\State Return VLEU score
\end{algorithmic}
\end{algorithm*}

\section{Human Evaluation}
\label{sec:human}

To demonstrate the effectiveness our VLEU metric evaluation, we conducted a comprehensive human evaluation study involving 10 human evaluators. These evaluators were tasked with creating a variety of T2I prompts, either freely or focused on specific subjects, similar to the prompt generation process of LLMs. The prompts generated by the evaluators were then used to produce images using the four T2I models under investigation.

\begin{figure}[h]
    \centering
    \includegraphics[width=\columnwidth]{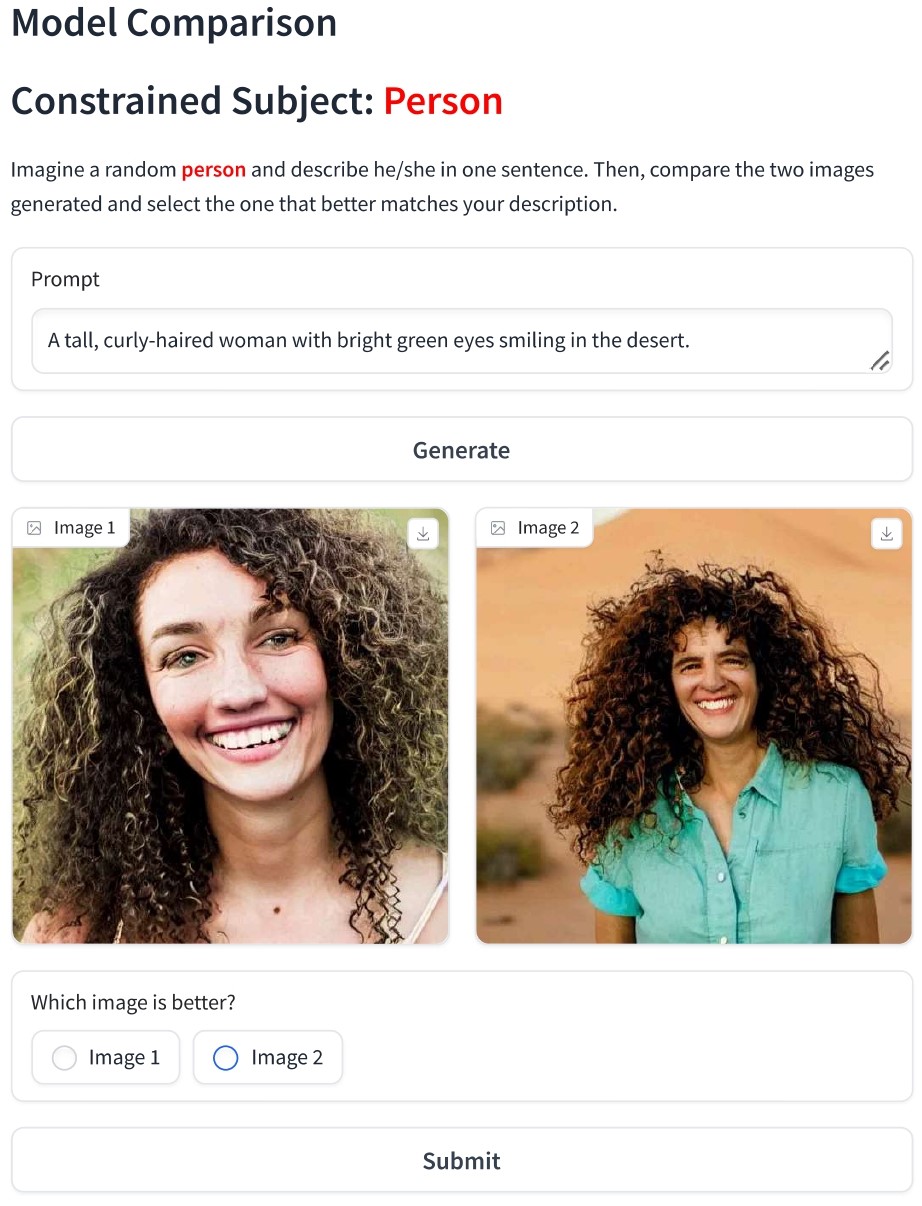}
    \caption{\textbf{Gradio interface used for human evaluation.} Evaluators first input a prompt related to a given subject based on the provided instructions, then click “Generate”. They then choose the better image from the two generated by different T2I models and submit their selection.}
    \label{fig:gradio_interface}
\end{figure}

To facilitate the evaluation process, we developed an interactive web interface using Gradio, which allowed evaluators to compare pairs of images generated by randomly selected pairs of models. Figure \ref{fig:gradio_interface} shows a screenshot of the Gradio interface used in the study. Evaluators were asked to determine which image better adhered to the given prompts, without knowing which model produced which image.

The pairwise comparisons made by the evaluators were used to compute an Elo rating for each model. The Elo rating system, originally developed for ranking chess players, is a method for calculating the relative skill levels of players in win-loss games. Each model's initial rating was set to 1000. The Elo rating for a model is updated based on the outcome of each pairwise comparison, with the winning model gaining points and the losing model losing points. The amount of points exchanged depends on the difference in the ratings of the two models, with larger differences resulting in smaller point exchanges.

The Elo rating \( R \) for a model is updated using the following formula:

\[
R_{\text{new}} = R_{\text{old}} + K \times (S - E)
\]

where:
- \( R_{\text{new}} \) is the new Elo rating.
- \( R_{\text{old}} \) is the old Elo rating.
- \( K \) is a constant that determines the sensitivity of the rating system (commonly set to 32).
- \( S \) is the actual score of the match (1 for a win, 0.5 for a draw, and 0 for a loss).
- \( E \) is the expected score, calculated using the formula:

\[
E = \frac{1}{1 + 10^{(R_{\text{opponent}} - R_{\text{old}})/400}}
\]

where \( R_{\text{opponent}} \) is the Elo rating of the opposing model.

\section{Hyperparameter Analysis}

In order to investigate the effectiveness of the VLEU metric concerning the number of sampled T2I prompts, we conducted multiple experiments across four T2I models. Each experiment involved resampling a certain number of T2I prompts using GPT 3.5 and computing the VLEU for each T2I model. The results, as shown in Table \ref{table:num}, indicate that as the number of sampled prompts increases, the VLEU score tends to rise. Moreover, even with sampling 1000 prompts, there is still a certain level of fluctuation in the VLEU obtained from two separate samplings. However, for a given set of sampled prompts, the relative ranking of VLEU scores among the models remains consistent. This suggests that while the absolute values of VLEU may vary between samplings, its ability to facilitate comparisons remains stable. Therefore, the choice of the number of prompts sampled could be made based on computational constraints or desired evaluation granularity, without significantly affecting the relative assessment of T2I model performance.

\section{Prompt Diversity}

\begin{figure*}
        \centering
        \begin{subfigure}{0.48\textwidth}
            \includegraphics[width=\linewidth]{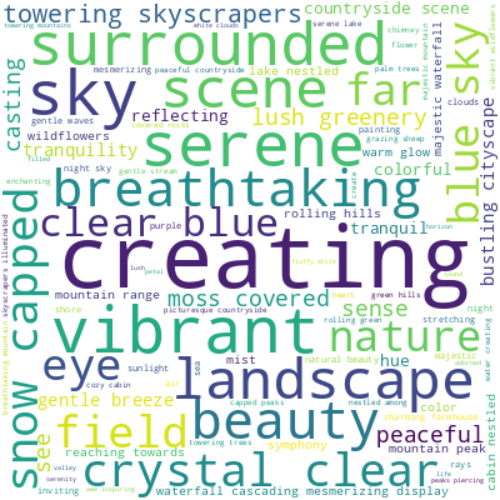}
            \subcaption{ChatGPT 3.5}
        \end{subfigure}%
        \hfill  %
        \begin{subfigure}{0.48\textwidth}
            \includegraphics[width=\linewidth]{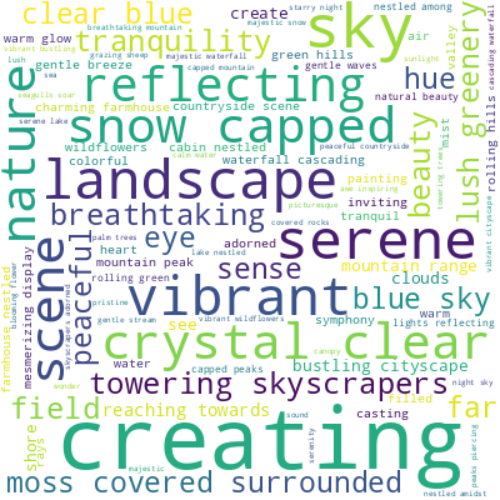}
            \subcaption{GPT-4}
        \end{subfigure}\\
        \begin{subfigure}{0.48\textwidth}
            \includegraphics[width=\linewidth]{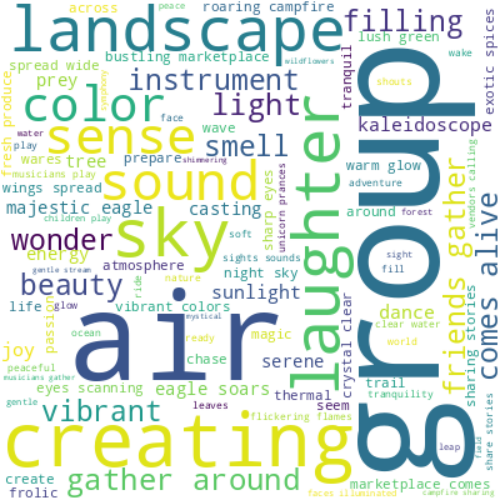}
            \subcaption{LLaMA-2-7B}
        \end{subfigure}%
        \hfill  %
        \begin{subfigure}{0.48\textwidth}
            \includegraphics[width=\linewidth]{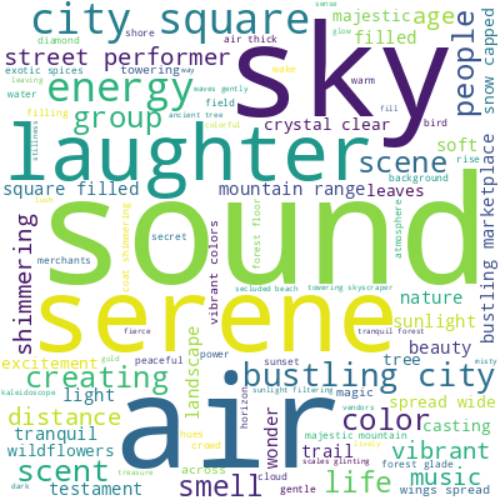}
            \subcaption{LLaMA-2-13B}
        \end{subfigure}
        \caption{\textbf{Word clouds of sampled prompts.} Each word cloud is generated from 1000 prompts, which were sampled by different LLMs on unconstrained subjects.}
        \label{fig:wordcloud}
\end{figure*}

To investigate the diversity of prompts generated by different LLMs, we generated word clouds for GPT-3.5, GPT-4, LLaMA-2-7B, and LLaMA-2-13B based on 1000 unconstrained prompts sampled from each model. As depicted in Figure \ref{fig:wordcloud}, the word clouds reveal notable differences in vocabulary richness and diversity among the prompts generated by these models.

From the word clouds, it is evident that prompts sampled from GPT-4 and GPT-3.5 exhibit a richer and more diverse vocabulary. In these word clouds, there is a greater variety of words, and no single word overly dominates, indicating a balanced and broad coverage of themes and subjects. On the other hand, the word clouds generated from LLaMA-2-7B and LLaMA-2-13B display a limited vocabulary with a few high-frequency words occupying a significant portion of the visual space. This suggests that these models produce less diverse prompts.

We observed that utilizing LLMs with better generalizability, such as GPT-4, as the Visual Text Sampler, resulted in higher VLEU scores compared to those obtained with LLMs with poorer generalizability, such as LLaMA-2-7B. This indicates that prompts from models like GPT-4 describe the images with more detailed and varied language, leading to a higher alignment with the generated images, thereby reflecting the model's generalizability more accurately. In contrast, LLMs with weaker descriptive capabilities produce prompts with substantially lower semantic alignment with the generated images, increasing the KL divergence and yielding higher VLEU scores. The generated word clouds and observed correlation between prompt diversity and VLEU scores underscore the importance of diverse and rich prompt generation in evaluating T2I models’ generalizability.

\end{document}